\documentclass[sigconf]{acmart}
\AtBeginDocument{%
  }

\setcopyright{acmlicensed}
\copyrightyear{2026}
\acmYear{2026}
\setcopyright{cc}
\setcctype{by}
\acmConference[KDD '26]{Proceedings of the 32nd ACM SIGKDD Conference on Knowledge Discovery and Data Mining V.2}{August 09--13, 2026}{Jeju Island, Republic of Korea}
\acmBooktitle{Proceedings of the 32nd ACM SIGKDD Conference on Knowledge Discovery and Data Mining V.2 (KDD '26), August 09--13, 2026, Jeju Island, Republic of Korea}
\acmDOI{10.1145/3770855.3819001}
\acmISBN{979-8-4007-2259-2/2026/08}

\usepackage{booktabs}
\usepackage{makecell}
\usepackage{multirow}
\usepackage{longtable}
\usepackage{balance}

\begin{document}

% \title{\textsc{Compass\scalebox{0.7}{\emoji{compass}}}: Navigating Global Marine Lead Data Integration through Expert-Guided LLM Agent}

\title{\textsc{Compass}: Navigating Global Marine Lead Data Integration through Expert-Guided LLM Agent}

\author{Yiming Liu}
\orcid{0009-0000-1893-8855}
\affiliation{
  \institution{School of Information Science and Electronic Engineering,\\Shanghai Jiao Tong University}
  \city{Shanghai}
  \country{China}}
\email{liu-ym@sjtu.edu.cn}

\author{Bin Lu}
\orcid{0000-0001-6452-7029}
\affiliation{%
  \institution{School of Information Science and Electronic Engineering,\\Shanghai Jiao Tong University}
  \city{Shanghai}
  \country{China}}
\email{robinlu1209@sjtu.edu.cn}
\authornote{Bin Lu and Meng Jin are co-corresponding authors.}

\author{Meng Jin}
\orcid{0000-0001-5960-4659}
\affiliation{%
  \institution{School of Artificial Intelligence,\\Shanghai Jiao Tong University}
  \city{Shanghai}
  \country{China}}
\email{jinm@sjtu.edu.cn}
\authornotemark[1]

\author{Ziyuan Sang}
\orcid{0009-0006-3091-4569}
\affiliation{
  \institution{School of Information Science and Electronic Engineering,\\Shanghai Jiao Tong University}
  \city{Shanghai}
  \country{China}
}
\email{sangziyuan@sjtu.edu.cn}

\author{Shuo Jiang}
\orcid{0000-0001-7860-1752}
\affiliation{
  \institution{State Key Laboratory of Estuarine and Coastal Research,\\East China Normal University}
  \city{Shanghai}
  \country{China}
}
\email{jiangshuo@sklec.ecnu.edu.cn}

\author{Lei Zhou}
\orcid{0000-0002-0433-3991}
\affiliation{
  \institution{School of Oceanography,\\Shanghai Jiao Tong University}
  \city{Shanghai}
  \country{China}
}
\email{zhoulei1588@sjtu.edu.cn}

\author{Xinbing Wang}
\orcid{0000-0002-0357-8356}
\affiliation{
  \institution{School of Information Science and Electronic Engineering,\\Shanghai Jiao Tong University}
  \city{Shanghai}
  \country{China}
  }
\email{xwang8@sjtu.edu.cn}

\author{Chenghu Zhou}
\orcid{0000-0003-3331-2302}
\affiliation{
  \institution{Institute of Geographical Science and Natural Resources Research,\\Chinese Academy of Sciences}
  \city{Beijing}
  \country{China}
  }
\email{zhouch@lreis.ac.cn}

\author{Jing Zhang}
\orcid{0000-0001-5403-5442}
\affiliation{
  \institution{State Key Laboratory of Estuarine and Coastal Research,\\East China Normal University}
  \city{Shanghai}
  \country{China}
}
\email{jzhang@sklec.ecnu.edu.cn}

\renewcommand{\shortauthors}{Yiming Liu et al.}

\begin{abstract}
Marine lead (Pb) and its isotopes are critical tracers for ocean circulation and anthropogenic pollution, yet in-situ observations remain costly and sparse. 
While vast historical records exist, they lie buried within the unstructured content of academic papers, creating ``data silos'' inaccessible to comprehensive analysis.
Manual extraction is unscalable, while general-purpose Large Language Models (LLMs) lack the necessary domain-specific knowledge, leading to hallucinations and scientifically invalid outputs.
To address this, we introduce an expert-guided adaptation approach that enables LLMs to perform rigorous scientific data extraction without fine-tuning. 
We operationalize this approach through \textsc{Compass}, an LLM agent framework enhanced by a Knowledge Tree co-designed with marine scientists, which decomposes complex tasks into verifiable steps, guiding the agent's reasoning to ensure scientific validity.
Deploying \textsc{Compass} across a corpus of over 230,000 relevant open-access papers, we successfully extract \textit{3,751 previously unincorporated Pb records}. This effort establishes the largest integrated marine Pb database to date. Beyond standard metrics, \textsc{Compass} demonstrates superior reliability through multi-layered validation, achieving 92\% accuracy as confirmed through expert manual verification. The newly integrated data expand coverage in previously under-sampled regions such as the East China Sea and the Southern Ocean, providing an enriched data foundation for future scientific discoveries. 
We release an interactive visualization platform\footnote{Please visit our online platform at \url{https://jingwei.acemap.cn/lead}.} to facilitate open scientific access. Our work demonstrates that expert-guided agents can effectively bridge the gap between general-purpose LLMs and high-stakes scientific domains, enabling scalable data discovery in geosciences.\footnote{Our code is available at \url{https://github.com/liuyiming01/COMPASS}.} 
\end{abstract}




\maketitle

\section{Introduction}
Trace elements and their isotopes (TEIs) in the ocean serve as vital diagnostic tools for deciphering Earth system dynamics. Among them, lead (Pb) and its isotopes are particularly important, functioning as powerful tracers for ocean circulation pathways and anthropogenic pollution history~\cite{HENDERSON2002257, boyle2014anthropogenic}. Since Patterson's seminal work~\cite{patterson1965contaminated} revealed global industrial Pb contamination, marine Pb has been extensively monitored to capture the temporal evolution of anthropogenic emissions.

However, the inherent scarcity of high-quality marine Pb data creates a significant bottleneck for global-scale analysis. Obtaining these records requires expensive oceanographic cruises and strict trace-metal-clean protocols~\cite{griffiths2020evaluation,ZHANG2024103212}, making every data point a high-value asset. While a substantial volume of historical in-situ observations has accumulated over decades, the vast majority remain buried within the unstructured text, tables, and figures of academic papers. This phenomenon has created a ``data silo'' problem, where critical datasets are scattered and effectively unavailable for global-scale synthesis.

\textbf{Prior works.} Existing efforts to bridge this gap have primarily relied on two integration paradigms: manual curation and automated extraction. Manual curation, exemplified by the GEOTRACES program~\cite{anderson2020geotraces} and analogous synthesis efforts across paleobiology~\cite{fan2020high} and terrestrial biogeochemistry~\cite{almaraz2025deep,li2024land}, ensures high data fidelity but is labor-intensive and unscalable against the exponential growth of literature. Conversely, automated approaches—spanning discriminative models~\cite{beltagy-etal-2019-scibert}, domain-specific LLMs~\cite{deng2024k2,bi-etal-2024-oceangpt}, multimodal document parsing~\cite{wang2024mineruopensourcesolutionprecise,zhao2024tabpedia}, and retrieval-augmented generation (RAG)~\cite{lewis2020retrieval,lala2023paperqa}—offer scalability but often treat scientific extraction as a probabilistic text task. Lacking domain-specific logical constraints, these methods frequently misinterpret fine-grained details—such as isotope notations or unit conversions—resulting in hallucinations that compromise scientific validity.

\textbf{Challenges.} To achieve both scalability and scientific rigor, effective automation requires addressing two fundamental barriers: (1) \textit{The Domain Knowledge Gap:} General-purpose LLMs lack the specialized knowledge and strict logical constraints required for scientific research. Without domain-specific guidance, these models often hallucinate or fail to distinguish between subtle scientific nuances, rendering their outputs unreliable for high-stakes analysis. (2) \textit{Structural Heterogeneity:} Scientific records are deeply embedded within unstructured text and diverse table formats. Extracting this fine-grained data requires a sophisticated, systematic workflow capable of navigating these complex structures, rather than simple information retrieval.

\textbf{Our Work.} To address these challenges, we propose \textsc{Compass}, a Knowledge Tree-enhanced LLM Agent framework designed to operationalize an expert-guided adaptation. Instead of relying on black-box fine-tuning, \textsc{Compass} adopts an \textit{expert-in-the-loop} philosophy. We co-designed a domain-specific Knowledge Tree with marine scientists, encoding rigorous reasoning logic to guide the automated process. Leveraging this structured guidance, \textsc{Compass} decomposes the integration task into hierarchical subtasks across three phases—collection, extraction, and aggregation—with validation checks embedded at each step to ensure that every extracted record satisfies physical constraints. By deploying \textsc{Compass} on over 230,000 relevant open-access papers, we successfully recovered 3,751 previously unincorporated Pb records, demonstrating the framework's capability to bridge the gap between everyday language models and rigorous scientific demands.
The main contributions of this work are summarized as follows: 
\begin{itemize}
    \item We develop \textsc{Compass}, a Knowledge Tree-enhanced LLM Agent framework that hierarchically decomposes complex scientific workflows, enabling accurate extraction and integration of fine-grained data (text and tables) from heterogeneous academic sources.
    \item We demonstrate that expert-guided knowledge injection via a co-designed Knowledge Tree can effectively bridge the gap between general-purpose LLMs and high-stakes scientific domains without fine-tuning, providing a practical and low-cost alternative to domain-specific model training.
    \item We deploy \textsc{Compass} to establish the largest integrated marine Pb database to date, recovering 3,751 previously unincorporated records from over 230,000 open-access papers. The integrated data substantially expand coverage in under-sampled regions such as the East China Sea and the Southern Ocean, providing an enriched data foundation for future scientific discoveries.
\end{itemize}

\begin{figure}[t]
  \centering
  \includegraphics[width=\linewidth]{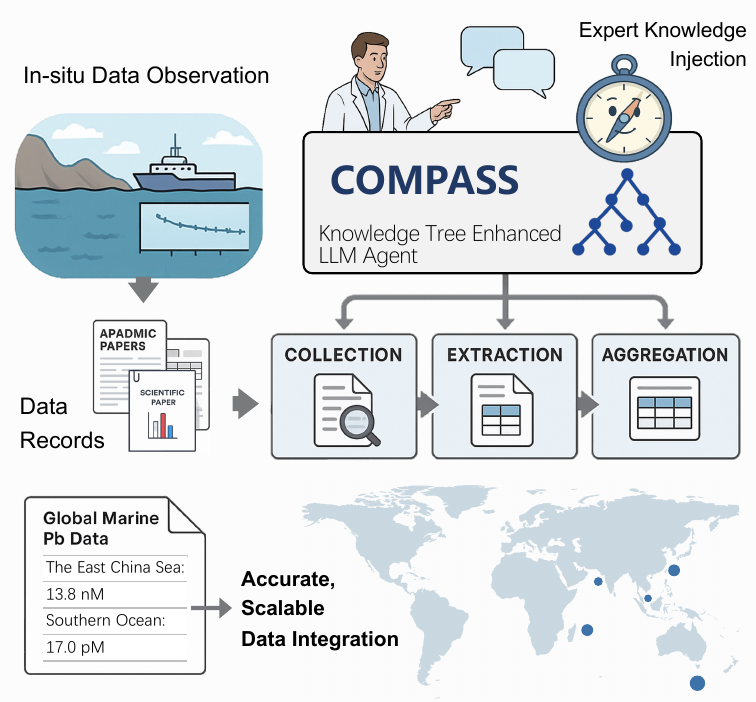}
  \caption{Overview of the \textsc{Compass} framework for marine Pb data integration.}
  \Description{A workflow diagram showing in-situ ocean observation and expert knowledge feeding into the COMPASS agent. Below, three sequential stages—Collection, Extraction, and Aggregation—process academic papers into integrated marine Pb data, visualized as a world map with scattered data points.}
  \label{fig:Intro}
\end{figure}

\begin{figure*}[t]
  \centering
  \includegraphics[width=\textwidth]{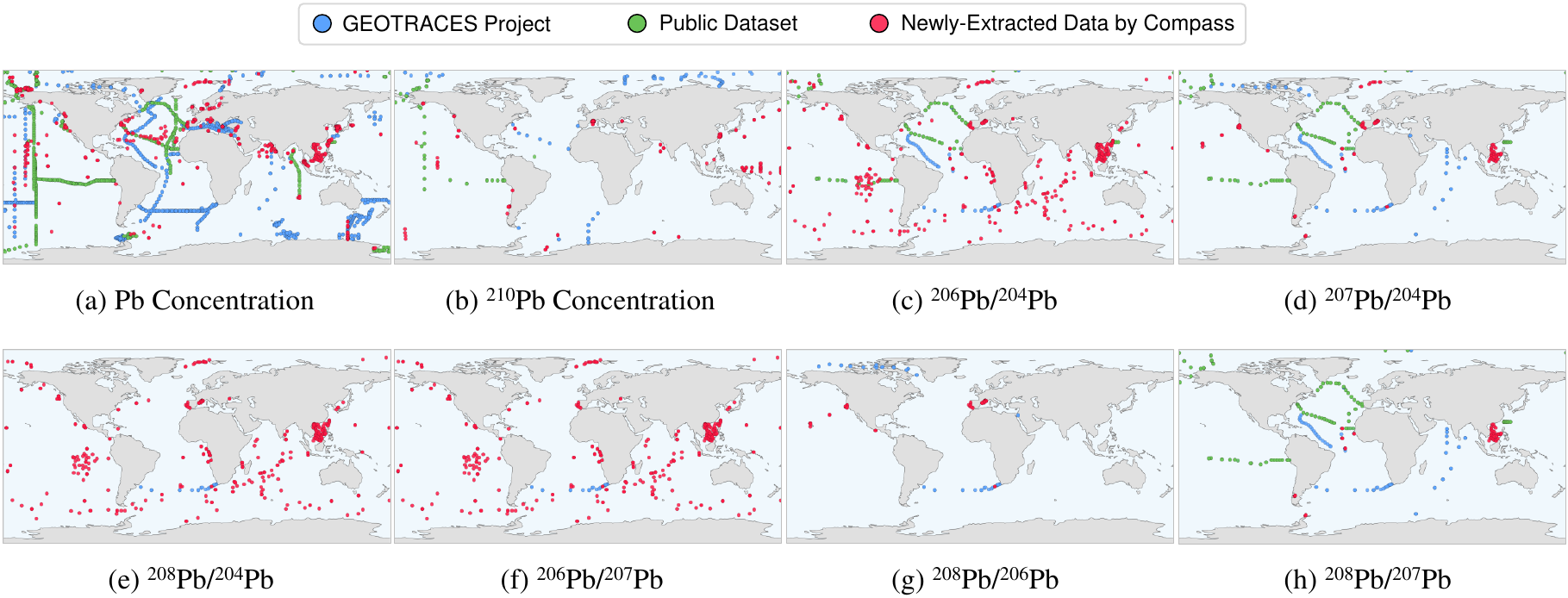}
  \caption{Global distribution of integrated marine Pb data showing improved coverage across different ocean regions. The eight panels represent different Pb measurements: (a) Pb concentration, (b) $^{210}$Pb concentration, and (c-h) six Pb isotope ratios. The map displays a total of 35,563 marine Pb records, including data extracted by \textsc{Compass} and from existing datasets, with color coding indicating data sources.}
  \Description{Eight world maps in a 2-by-4 grid, each showing colored dots for different Pb measurements across oceans. Blue dots represent GEOTRACES data, green for public datasets, and red for newly extracted Compass data, with dense red clusters in the East China Sea, Arabian Sea, and Southern Ocean.}
  \label{fig:Pb_map}
\end{figure*}

\section{Related Work}

\subsection{Scientific Data Integration}
Integrating heterogeneous scientific data is a cornerstone of data-driven discovery. Existing approaches fall into two paradigms.
\textit{(1) Manual Integration.}
Research programs such as GEOTRACES~\cite{anderson2020geotraces} and regional syntheses~\cite{zurbrick2017historic} enable systematic integration of marine trace element data through international collaboration.
Fan et al.~\cite{fan2020high} manually synthesize data from thousands of marine fossil species to generate high-resolution biodiversity curves.
Similar efforts in terrestrial biogeochemical cycles~\cite{almaraz2025deep,li2024land,li2021terrestrial} yield important insights but require substantial human effort, becoming increasingly impractical with the exponential growth of publications.
\textit{(2) Automated Integration.}
Text-based extraction methods have evolved from discriminative models like SciBERT~\cite{beltagy-etal-2019-scibert} to structured extraction with domain-specific and multi-expert language models~\cite{dagdelen2024structured,10.1145/3690624.3709400,10.1145/3637528.3671975}.
Cross-modality frameworks such as TabPedia~\cite{zhao2024tabpedia}, MinerU~\cite{wang2024mineruopensourcesolutionprecise}, MatViX~\cite{khalighinejad-etal-2025-matvix}, and Uni-SMART~\cite{cai2024uni} extend beyond text to support joint analysis of tables and figures, with applications spanning document extraction to clinical and materials science data integration~\cite{lipkova2022artificial, bazgir2025multicrossmodal}.
Despite this progress, existing automated methods primarily focus on information extraction from individual documents rather than end-to-end scientific synthesis with physical constraint validation.

\subsection{Scientific LLM Agents}
LLM agents leverage advanced comprehension and reasoning to automate scientific workflows.
\textit{(1) Training-based Agents.}
Domain-specific LLMs such as K2~\cite{deng2024k2} for geoscience and OceanGPT~\cite{bi-etal-2024-oceangpt} for ocean science improve domain knowledge through curated corpora and supervised training.
Agent frameworks further employ curriculum learning and reinforcement learning—for example, AutoWebGLM~\cite{lai2024autowebglm} integrates both to improve web navigation for scientific tasks.
Multi-agent strategies~\cite{10.1145/3690624.3709205,10.1145/3580305.3599359,10.1145/3580305.3599379,10.1145/3580305.3599856,10.1145/3580305.3599241,10.1145/3580305.3599272} enable large-scale coordination across domains but typically demand significant computational resources and domain-specific datasets.
\textit{(2) Prompting and Retrieval-based Agents.}
PaperQA~\cite{lala2023paperqa} employs retrieval-augmented generation (RAG) to perform evidence-based information retrieval across scientific articles.
ChemAgent~\cite{tang2025chemagent} and SciAgents~\cite{https://doi.org/10.1002/adma.202413523} leverage tool-use and ontological knowledge graphs for multi-agent reasoning, while other approaches explore self-updating mechanisms and automated environment generation~\cite{10.1145/3690624.3709321, 10.1145/3690624.3709171}.
However, standard RAG or general agents lack structured domain logic for fine-grained integration in high-stakes scientific fields.
\textsc{Compass} addresses this gap through an expert-guided adaptation approach: instead of probability-based retrieval or costly fine-tuning, it embeds structured expert knowledge directly into the agent's workflow, ensuring scientific validity while maintaining scalability.

\section{Background and Problem Formulation}

\subsection{Scientific Significance of Marine Lead}

\begin{figure}[t]
  \centering
  \includegraphics[width=.9\linewidth]{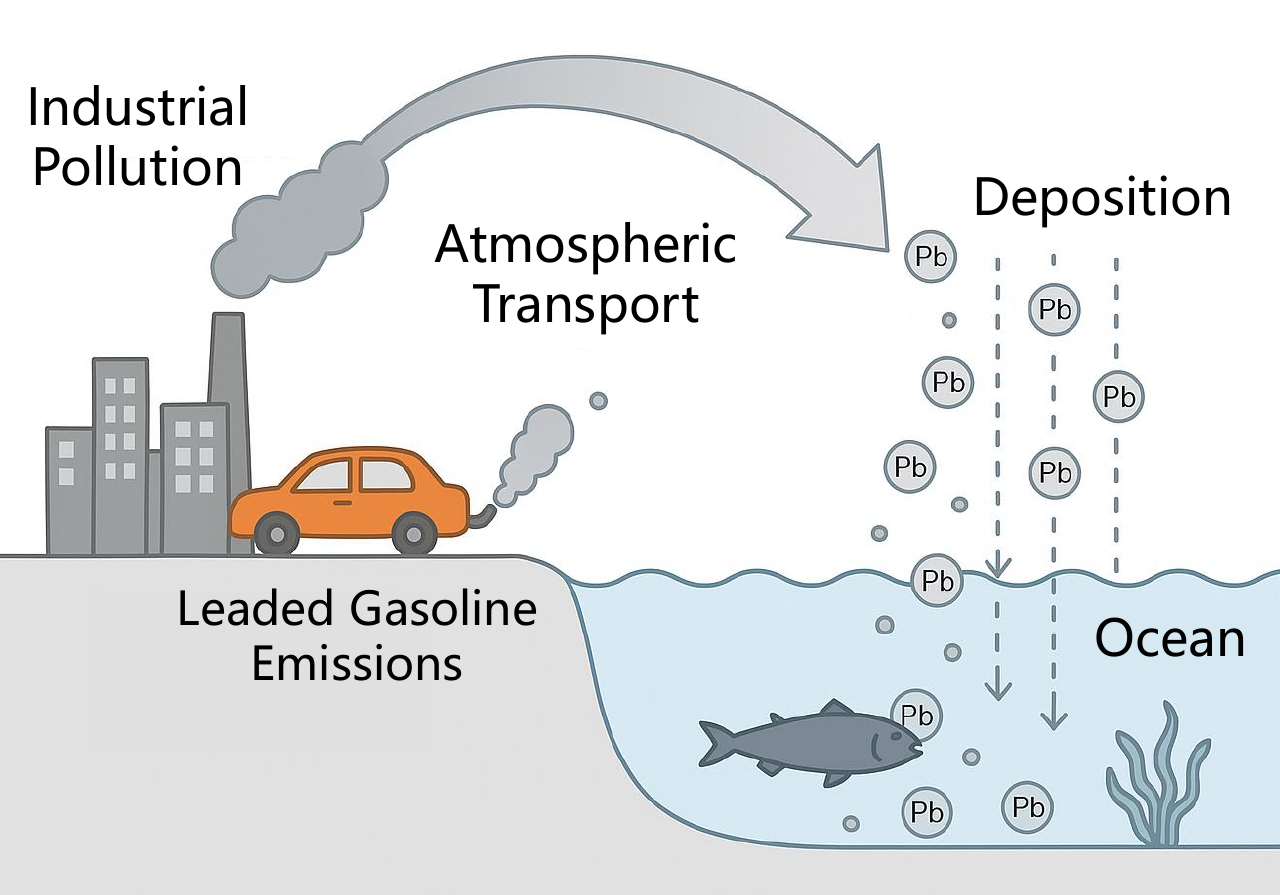}
  \caption{Pb from anthropogenic emissions enters the ocean through atmospheric transport and surface deposition.}
  \Description{A schematic showing industrial factories and a car emitting lead on the left, with a curved arrow labeled Atmospheric Transport carrying Pb atoms over the ocean, where they deposit into the water around a fish and seaweed.}
  \label{fig:Lead_transport}
\end{figure}

Marine lead (Pb) and its isotopes serve as critical tracers for ocean circulation pathways and anthropogenic pollution history~\cite{HENDERSON2002257, boyle2014anthropogenic}.
Pb primarily originates from anthropogenic activities such as leaded gasoline combustion and industrial emissions~\cite{nriagu1988quantitative}, dispersing via atmospheric transport to even remote ocean regions~\cite{cziczo2009inadvertent} (Figure~\ref{fig:Lead_transport}).
Its relatively brief oceanic residence time compared to global mixing timescales enables Pb to retain distinct regional isotopic fingerprints that record transient environmental changes~\cite{SCHAULE198197, WU19973279}, while obtaining high-quality measurements requires expensive oceanographic cruises and strict trace-metal-clean protocols~\cite{griffiths2020evaluation,ZHANG2024103212}.

However, individual studies provide only geographically limited snapshots. A comprehensive integrated database unlocks analyses no single study can achieve: global Pb inventory and distribution estimation, source identification via stable isotope ratios (e.g., $^{206}$Pb/$^{207}$Pb, $^{208}$Pb/$^{206}$Pb), and investigation of oceanic responses to the global phase-out of leaded gasoline. Multi-decadal observations, when unified, further support biogeochemical modeling and environmental policy evaluation.
Despite this potential, vast historical data remain fragmented across decades of publications. Manual curation programs like GEOTRACES~\cite{anderson2020geotraces} and regional syntheses~\cite{zurbrick2017historic} are fundamentally unscalable, directly motivating the integration framework presented in this work.

\subsection{The Data Integration Challenge}

\textbf{Data Source Categorization.}
Marine Pb data exists across three categories: \textit{structured databases} ($\mathcal{D}_{structured}$) such as GEOTRACES~\cite{anderson2020geotraces} with well-defined schemas; \textit{scattered datasets} ($\mathcal{D}_{scattered}$) from individual studies with varying conventions; and \textit{academic papers} ($\mathcal{D}_{papers}$), where fine-grained measurements are embedded in unstructured text and diverse table formats.

\textbf{Technical Challenges.}
While the general challenges of domain knowledge gap and structural heterogeneity apply broadly, marine Pb data integration presents additional domain-specific difficulties:
(1)~\textit{Data Source Proliferation:} Pb measurements scatter across literature spanning over five decades, making manual integration impractical.
(2)~\textit{Fine-grained Semantic Complexity:} Subtle distinctions—such as dissolved versus particulate phases, different isotope ratio conventions, and varied analytical methods—are easily confused by general-purpose models.
(3)~\textit{Cross-source Harmonization:} Inconsistent units, depth references, and coordinate formats demand systematic normalization.

\textbf{Problem Formulation.}
Each source exhibits distinct schema structures, semantic conventions, and quality standards (e.g., varying units, precision levels, and coordinate formats), necessitating a hierarchical pipeline that progressively resolves these heterogeneities.
We formalize the data integration objective as constructing a unified dataset $D^*$ from heterogeneous sources $\mathcal{D} = \{D_1, D_2, \ldots, D_n\}$ through a hierarchical processing pipeline:
\begin{align}
D^* = \mathcal{A}(\mathcal{E}(\mathcal{C}(\mathcal{D}_{papers})), \mathcal{D}_{structured}, \mathcal{D}_{scattered}),
\end{align}
where $\mathcal{C}$ denotes the collection function that identifies relevant papers from a large candidate corpus, $\mathcal{E}$ denotes the extraction function that resolves schema heterogeneity and retrieves structured records from paper tables, and $\mathcal{A}$ denotes the aggregation function that integrates all sources with standardized units, formats, and quality controls.
$\mathcal{D}_{papers} \subset \mathcal{D}$, $\mathcal{D}_{structured} \subset \mathcal{D}$, and $\mathcal{D}_{scattered} \subset \mathcal{D}$ represent the academic paper sources, structured database sources, and scattered dataset sources, respectively.
This formulation directly motivates the three-phase architecture of \textsc{Compass}.

\section{\textsc{Compass} Framework}

\subsection{Expert-Guided Adaptation Paradigm}

Applying LLMs to scientific data integration requires bridging the gap between general-purpose language understanding and rigorous domain reasoning. Existing strategies each have fundamental limitations for this goal.
\textit{Fine-tuning}~\cite{deng2024k2, bi-etal-2024-oceangpt} improves domain knowledge but incurs high computational cost and often degrades the instruction-following capabilities critical for multi-step workflows.
\textit{Retrieval-Augmented Generation (RAG)}~\cite{lewis2020retrieval} injects knowledge at inference time but treats it as unstructured text fragments, lacking mechanisms to enforce logical constraints or sequential validation across complex pipelines.
\textit{Knowledge graph (KG) and ontology-based approaches}~\cite{https://doi.org/10.1002/adma.202413523} organize domain concepts and semantic relationships, but primarily capture declarative knowledge (\textit{what is}) rather than procedural knowledge (\textit{how to act}).

\textsc{Compass} introduces an \textit{expert-guided adaptation paradigm} that encodes domain expertise as a structured, executable Knowledge Tree co-designed with domain scientists. Unlike the above approaches, the Knowledge Tree integrates operational logic—prescribing not only \textit{what} the agent should know but also \textit{how} it should reason and validate its outputs—making the workflow interpretable, controllable, and scientifically trustworthy.

\subsection{Knowledge Tree: Design and Construction}

\textbf{Definition (Knowledge Tree).} A Knowledge Tree $T = (N, H, K)$ is a hierarchical structure where $N$ represents the set of nodes corresponding to specific tasks or subtasks, $H$ denotes directed edges representing parent-child decomposition relationships, and $K$ contains expert-provided knowledge constraints associated with each node. Each node $n_{i,j} \in N$ at level $L_i$ corresponds to subtask $t_{i,j}$ and can be decomposed into child nodes at level $L_{i+1}$, with leaf nodes representing atomic tasks requiring direct execution.

\textbf{Knowledge Structure.} For each node $n_{i,j}$, the associated knowledge $K_{i,j} = \{BK, LC, OG, VC\}$ encompasses four complementary dimensions:
\begin{itemize}
\item \textit{Background Knowledge} ($BK$): Domain concepts, terminology, and contextual information providing foundational understanding for task execution.
\item \textit{Logical Constraints} ($LC$): Explicit rules, conditional dependencies, and logical requirements defining valid execution paths.
\item \textit{Operational Guidelines} ($OG$): Step-by-step procedures, decision heuristics, and output format specifications ensuring standardized execution.
\item \textit{Validation Criteria} ($VC$): Quality metrics, physical constraint checks, and verification steps enabling automated assessment of output correctness.
\end{itemize}
This four-dimensional structure ensures that each node carries not only declarative knowledge but also procedural and evaluative guidance, distinguishing our approach from flat structured prompting or static ontologies.

\textbf{Construction Protocol.} The Knowledge Tree is constructed offline by AI researchers in collaboration with domain experts, following a four-step protocol for each hierarchical level $L_i$:
\begin{enumerate}
    \item \textit{Task Context Initialization:} AI researchers present each subtask along with its hierarchical dependencies, providing domain experts with comprehensive understanding of the task scope and requirements.
    \item \textit{Expert Knowledge Elicitation:} Domain experts provide specialized knowledge across the four dimensions ($BK$, $LC$, $OG$, $VC$) based on their domain expertise and practical experience.
    \item \textit{Knowledge Structuring:} AI researchers organize expert-provided knowledge into structured knowledge nodes, ensuring consistency for subsequent automated processing.
    \item \textit{Completeness Validation:} Domain experts review the constructed knowledge structure for comprehensive coverage of domain requirements before proceeding to the next level.
\end{enumerate}
This four-step protocol is applied iteratively for each hierarchical level, with successive rounds refining the knowledge structure based on expert feedback and agent execution results on representative samples.
The finalized Knowledge Tree is then integrated into \textsc{Compass} to guide agent deployment.

\subsection{Marine Pb Knowledge Tree}
\label{sec:compass}
\begin{figure}[t]
  \centering
  \includegraphics[width=\linewidth]{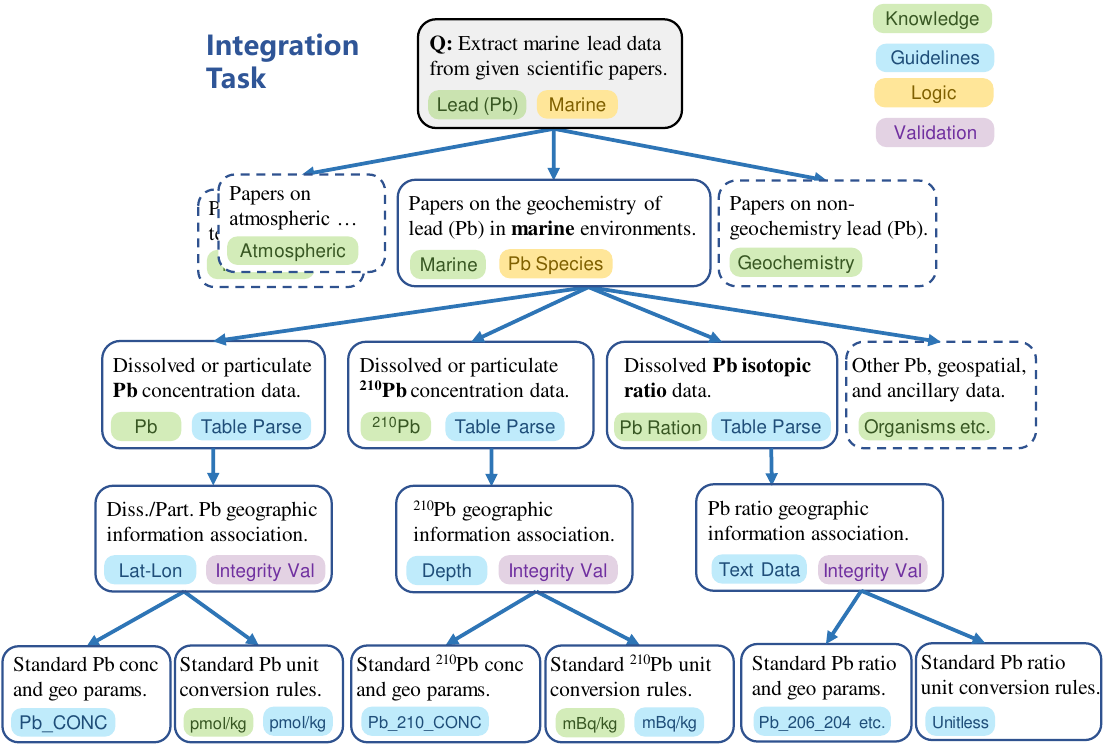}
  \caption{Domain Knowledge Tree for marine Pb data integration, co-designed by AI researchers and marine scientists.}
  \Description{A hierarchical tree diagram with a root Integration Task node branching into three categories—Atmospheric, Marine Pb Species, and Geochemistry—each further decomposing into specific data types and standardization rules at leaf nodes.}
  \label{fig:Pb_tree}
\end{figure}

We collaborate with marine scientists to construct the Knowledge Tree for marine Pb data integration.
Figure~\ref{fig:Pb_tree} shows the resulting structure. The overall task is decomposed into hierarchical subtasks, with Pb-related papers classified into fine-grained categories for accurate identification.
Marine Pb data encompasses multiple types—including Pb concentrations (dissolved, particulate), $^{210}$Pb activity concentrations, and six Pb isotope ratios—each requiring specific data formats, transformation logic, and validation standards.
For each subproblem, marine scientists provide the corresponding $BK$, $LC$, $OG$, and $VC$ knowledge. 

In practice, the Knowledge Tree construction is a one-time effort: the marine Pb tree (approximately 20 nodes) was built in 6--7 hours by two marine scientists in collaboration with AI researchers.

\begin{figure}[t]
  \centering
  \includegraphics[width=\linewidth]{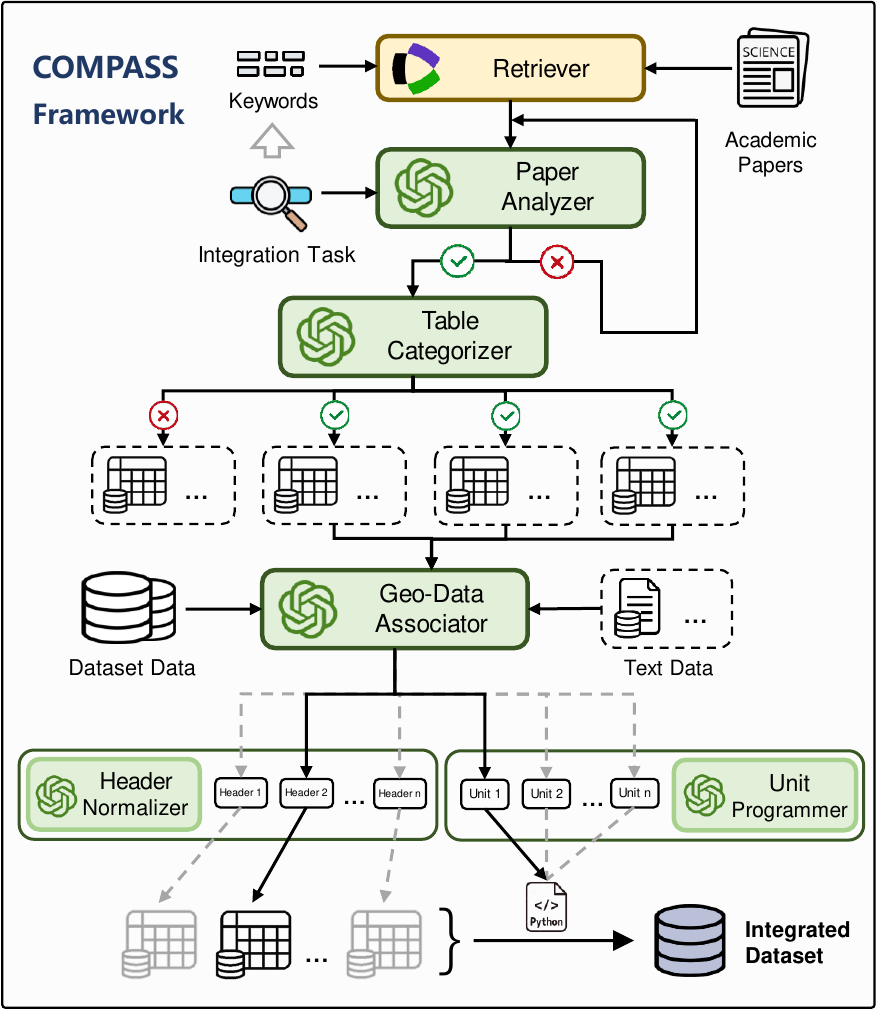}
  \caption{Architecture of \textsc{Compass}: hierarchical agent components and their interactions. (top-down view)}
  \Description{A vertical pipeline diagram showing Keywords feeding into a Retriever, then Paper Analyzer, Table Categorizer with check and cross marks, Geo-Data Associator merging tables with dataset and text data, and finally Header Normalizer and Unit Programmer converging into an Integrated Dataset.}
  \label{fig:Compass}
\end{figure}

\subsection{Agent Deployment}
\label{sec:agent_deployment}
When LLMs directly process complex data integration tasks, they often produce incomplete execution and logical inconsistencies, struggling to maintain coherent reasoning across interdependent subtasks.
Inspired by the analytic hierarchy process~\cite{vaidya2006analytic} and hierarchical decision trees~\cite{hunt1966experiments}, \textsc{Compass} systematically decomposes complex tasks into a three-phase hierarchical pipeline with five specialized components, as shown in Figure~\ref{fig:Compass}.

\textbf{(1) Collection Phase.} The \textit{paper classification} component retrieves candidate papers from academic repositories (e.g., Semantic Scholar~\cite{DBLP:conf/naacl/AmmarGBBCDDEFHK18}, AceMap~\cite{DBLP:journals/corr/abs-2403-02576}) via keyword search, then applies LLM-powered semantic analysis guided by the Knowledge Tree to identify papers containing target data.

\textbf{(2) Extraction Phase.} The \textit{table classification} component integrates the PDF parsing tool MinerU~\cite{wang2024mineruopensourcesolutionprecise} to extract tabular content, then leverages LLMs to analyze tables with contextual information (captions, in-text descriptions) to classify and filter target data tables.

\textbf{(3) Aggregation Phase.} Three components handle different integration aspects: the \textit{data association} component links related elements within papers (e.g., matching Pb measurements with sampling coordinates from separate tables); the \textit{cross-source fusion} component normalizes heterogeneous table headers into a unified schema; and the \textit{data standardization} component converts values into consistent units through LLM-generated conversion functions.

For each component, the corresponding Knowledge Tree node $K_{i,j}$ is translated into a domain-aware prompt that systematically integrates all four knowledge dimensions ($BK$, $LC$, $OG$, $VC$); the complete prompts are publicly available in our GitHub repository, ensuring that the LLM operates with domain-specific context, constrained reasoning boundaries, standardized output formats, and built-in quality checks. A record of completed subtasks and their outputs is maintained across hierarchical levels to preserve consistency. While this study focuses on text and tables, the modular architecture is designed to accommodate additional tools as supplementary components in future extensions. Formally, \textsc{Compass} operates as a state transition system where each completed subtask represents a distinct state, with transitions governed by the corresponding Knowledge Tree constraints, ensuring that domain expertise directs every decision point while maintaining systematic progression through the task hierarchy.

\subsection{Validation Mechanism}
\label{sec:validation}

Scientific data integration demands rigorous quality assurance. \textsc{Compass} implements a multi-layered validation mechanism informed by the Validation Criteria ($VC$) in the Knowledge Tree.

\textbf{Automated Physical Constraint Checks.} Every extracted record undergoes value range verification against physically plausible bounds for each data type, unit conversion cross-validation, and geographical outlier filtering against ocean boundary definitions. When validation failures are detected, \textsc{Compass} rolls back the current step and re-processes the failed subtask, preventing error propagation to downstream components.

\textbf{Expert Manual Validation.} Beyond automated checks, we conduct manual validation with marine scientists on a randomly selected subset, providing an independent assessment of end-to-end accuracy. Results are reported in Section~\ref{sec:data_quality}.

\section{Experiments and Results}

\subsection{Pre-Deployment Evaluation}

Before full-scale deployment, we conduct a comprehensive evaluation of \textsc{Compass} to validate its accuracy on marine Pb data integration. 
We present detailed experimental analyses aimed at addressing the following three research questions (\textbf{RQs}):

\begin{itemize}
    \item \textbf{RQ1:} Can \textsc{Compass} accurately integrate target marine Pb data, and how does it compare with general-purpose LLMs?
    \item \textbf{RQ2:} What are the advantages of \textsc{Compass} over domain-specific fine-tuned LLMs in terms of performance and cost?
    \item \textbf{RQ3:} How does each core component contribute to overall performance?
    % \item \textbf{RQ4:} What are the limitations of LLM-based agent \textsc{Compass} and possible directions for improvement?
\end{itemize}

\begin{table}[t]
    \centering
    \caption{Categories of academic papers in the marine Pb benchmark dataset.}
    \label{tab:paper_categories}
    \resizebox{\columnwidth}{!}{
    \begin{tabular}{ll}
        \toprule
        \textbf{Category} & \textbf{Description} \\
        \midrule
        Marine Pb conc.         & Papers reporting dissolved or particulate Pb\\
                               & concentrations in marine environments. \\
        Marine $^{210}$Pb      & Papers reporting $^{210}$Pb concentrations in \\
                               & marine environments. \\
        Marine Pb isotopes ratios & Papers reporting Pb isotope ratios \\
                               & (e.g., $^{206}$Pb/$^{204}$Pb) in marine environments. \\
        Marine Pb (non-target) & Papers containing marine Pb data irrelevant \\
                               & to target tasks (negative control). \\
        Atmospheric Pb         & Papers focusing on atmospheric Pb sources \\
                               & and deposition processes. \\
        Terrestrial Pb         & Papers focusing on Pb in terrestrial systems \\
                               & (soils, freshwater, etc.). \\
        Analytical Pb          & Papers on laboratory methods and analytical \\
                               & techniques for Pb analysis. \\
        Irrelevant ``Pb''        & Papers mentioning ``Pb'' but unrelated to \\
                               & lead or environmental science. \\
        Other marine elements  & Papers on marine geochemistry of elements \\
                               & other than Pb. \\
        Unrelated topics       & Papers completely unrelated to Pb research \\
                               & or marine science. \\
        \bottomrule
    \end{tabular}
    }
\end{table}

\subsubsection{Experimental Setup}

We define the evaluation task, describe the benchmark dataset, and specify the metrics for evaluation.

\textbf{Task Overview.}  
To evaluate whether \textsc{Compass} can accurately perform scientific data integration, we structure the evaluation into three core experimental tasks. This stepwise design mirrors the typical workflow used by researchers:

\begin{itemize}
    \item \textit{Paper Classification:} Identify whether a given paper belongs to the specific subfield relevant to the data integration task.
    \item \textit{Table Classification:} Identify whether a specific table within a paper includes target data for extraction.
    \item \textit{End-to-End Extraction:} Extract all target data points from the entire paper collection.
\end{itemize}

\textbf{Benchmark Dataset.}  
To evaluate extraction performance, we construct a benchmark dataset distributed across 10 categories (Table~\ref{tab:paper_categories}). ``Marine Pb'', ``Atmospheric Pb'', and ``Terrestrial Pb'' are used for the Paper Classification task, as these reflect key environmental compartments relevant to Pb research. Within Marine Pb, we distinguish three target subcategories—Marine Pb conc., Marine $^{210}$Pb, and Marine Pb isotope ratios—whose associated tables provide the target data for table classification and end-to-end data extraction tasks. Tables containing target marine Pb data serve as positive examples, while tables from the remaining seven categories act as negative controls to assess classification robustness. We curated a total of 337 tables, among which 63 contain target marine Pb data, yielding 1,397 data points for downstream extraction tasks.

\textbf{Metrics and Configuration.}
We employ accuracy (Acc), precision (Prec), recall (Rec), and F1-score for comprehensive quantitative evaluation. All models use greedy decoding (temperature=0) for deterministic and reproducible results. Open-source LLMs are loaded with 4-bit NF4 quantization (BitsAndBytes) with fp16 inference. Maximum generation length is set to 32 tokens for classification and 2,048 tokens for extraction tasks. All experiments are conducted on two NVIDIA RTX 3090 GPUs (24GB VRAM each).

\textbf{Baselines.}
As there are no agent-based approaches for scientific data integration, we compare \textsc{Compass} with two categories of baselines.
\textit{General-purpose LLMs}: GPT-4o, Gemini-2.5-pro, Llama-3.1-8B-Instruct, Qwen3-8B, and Qwen2.5-32B-Instruct, representing state-of-the-art proprietary and open-source models.
\textit{Domain-specific LLMs}: K2~\cite{deng2024k2}, a Llama-based model pretrained on geoscience data (limited to 2,048-token context), and OceanGPT~\cite{bi-etal-2024-oceangpt} (Qwen2-7B-based), fine-tuned on ocean science data.
To ensure fairness, all baselines are evaluated using identical, iteratively refined prompts (available in our repository).

\subsubsection{Results and Analysis}

\begin{table}[t]
    \caption{Performance comparison on the marine Pb benchmark.}
    \label{tab:main_result}
    \resizebox{\columnwidth}{!}{
    \begin{tabular}{lcccccccc}
        \toprule
        \textbf{Model} & \textbf{Size} & \multicolumn{2}{c}{\makecell{\textbf{Paper}\\\textbf{Classification}}}
                        & \multicolumn{2}{c}{\makecell{\textbf{Table}\\\textbf{Classification}}}
                        & \multicolumn{3}{c}{\makecell{\textbf{End-to-End}\\\textbf{Extraction}}} \\
    \cmidrule(lr){3-4} \cmidrule(lr){5-6} \cmidrule(lr){7-9}
    & & \textbf{Acc} & \textbf{F1} & \textbf{Acc} & \textbf{F1} & \textbf{Prec} & \textbf{Rec} & \textbf{F1} \\
        \midrule
        GPT-4o & N/A & 0.930 & 0.845 & 0.778 & 0.688 & 0.429 & 0.330 & 0.373 \\
        Gemini-2.5-pro & N/A & 0.960 & 0.913 & 0.724 & 0.681 & 0.334 & \textbf{0.511} & 0.404 \\
        K2 (LLaMA) & 7B & 0.050 & 0.037 & 0.000 & 0.000 & 0.000 & 0.000 & 0.000 \\
        OceanGPT & 7B & 0.800 & 0.748 & 0.312 & 0.385 & 0.038 & 0.005 & 0.009 \\
        Llama-3.1 & 8B & 0.820 & 0.757 & 0.691 & 0.613 & 0.067 & 0.085 & 0.075 \\
        Qwen3 & 8B & 0.880 & 0.801 & 0.769 & 0.658 & 0.175 & 0.223 & 0.196 \\
        Qwen2.5 & 32B & 0.920 & 0.887 & 0.766 & 0.684 & 0.320 & 0.394 & 0.353 \\
        \midrule
        \multirow{2}{*}{\textsc{Compass} (Ours)}
            & 8B  & 0.960 & 0.942 & \textbf{0.947} & \textbf{0.897} & 0.196 & 0.158 & 0.175 \\
            & 32B & \textbf{0.960} & \textbf{0.956} & 0.920 & 0.865 & \textbf{0.508} & 0.429 & \textbf{0.465} \\
        \bottomrule
    \end{tabular}
    }
\end{table}

\textbf{Overall Performance (RQ1).} As shown in Table~\ref{tab:main_result}, \textsc{Compass} achieves the highest scores across all tasks, exceeding 90\% accuracy in both classification tasks. In end-to-end extraction, \textsc{Compass} (32B) achieves 0.465 F1, outperforming GPT-4o (0.373) and Gemini-2.5-pro (0.404). 
The advantage is most pronounced in table classification, where Knowledge Tree guidance enables accurate distinction between target Pb tables and scientifically 
irrelevant ones. While Gemini-2.5-pro achieves higher recall (0.511 vs.\ 0.429), its substantially lower precision (0.334 vs.\ 0.508) reflects a tradeoff unsuitable for scientific database construction, where data integrity is paramount.

\textbf{Advantages over Fine-tuned Models (RQ2).} Domain-specific models perform poorly despite their domain training: K2 fails due to context length limitations, while OceanGPT achieves only 0.009 F1 in extraction. Fine-tuning enhances domain vocabulary but degrades the instruction-following capabilities essential for multi-step integration. \textsc{Compass} preserves these capabilities while injecting expertise through the Knowledge Tree, avoiding costly retraining.

\textbf{Ablation Study (RQ3).} Table~\ref{tab:ablation} shows that removing tree structured logic reduces end-to-end F1 from 0.465 to 0.402, while removing knowledge nodes reduces it to 0.381. The tree structure ensures systematic task decomposition, while expert knowledge enables scientifically grounded decisions---their combination yields synergistic improvements. The higher recall but lower precision of the ablated variant reflects a precision--recall tradeoff: less structured extraction recovers more records at the cost of data integrity. Enabling automated validation rollback---which directly exercises the Validation Criteria ($VC$) dimension---further improves end-to-end F1 to 0.619, demonstrating $VC$'s contribution to overall reliability. Rollback results are reported here rather than in Table~\ref{tab:main_result}, as baselines are evaluated under single-pass extraction for fair comparison.

\begin{table}[t]
    \caption{Ablation study of \textsc{Compass} (Qwen2.5-32B backbone). ``+rollback'' activates the full pipeline with automated validation rollback, directly exercising the Validation Criteria ($VC$) knowledge dimension.}
    \label{tab:ablation}
    \resizebox{\columnwidth}{!}{
    \begin{tabular}{lccccccc}
        \toprule
        \textbf{Model} & \multicolumn{2}{c}{\makecell{\textbf{Paper}\\\textbf{Classification}}}
                        & \multicolumn{2}{c}{\makecell{\textbf{Table}\\\textbf{Classification}}}
                        & \multicolumn{3}{c}{\makecell{\textbf{End-to-End}\\\textbf{Extraction}}} \\
    \cmidrule(lr){2-3} \cmidrule(lr){4-5} \cmidrule(lr){6-8}
    & \textbf{Acc} & \textbf{F1} & \textbf{Acc} & \textbf{F1} & \textbf{Prec} & \textbf{Rec} & \textbf{F1} \\
        \midrule
        Qwen2.5-32B (vanilla) & 0.920 & 0.887 & 0.766 & 0.684 & 0.320 & 0.394 & 0.353 \\
        w/o tree-structured logic & 0.940 & 0.944 & 0.864 & 0.800 & 0.364 & 0.449 & 0.402 \\
        w/o knowledge nodes & 0.950 & 0.907 & 0.896 & 0.825 & 0.382 & 0.380 & 0.381 \\
        \midrule
        \textsc{Compass} (Ours) & 0.960 & 0.956 & 0.920 & 0.865 & 0.508 & 0.429 & 0.465 \\
        \textsc{Compass} + rollback & \textbf{0.960} & \textbf{0.956} & \textbf{0.920} & \textbf{0.865} & \textbf{0.750} & \textbf{0.526} & \textbf{0.619} \\
        \bottomrule
    \end{tabular}
    }
\end{table}

\begin{figure*}[t]
  \centering
  \includegraphics[width=\textwidth]{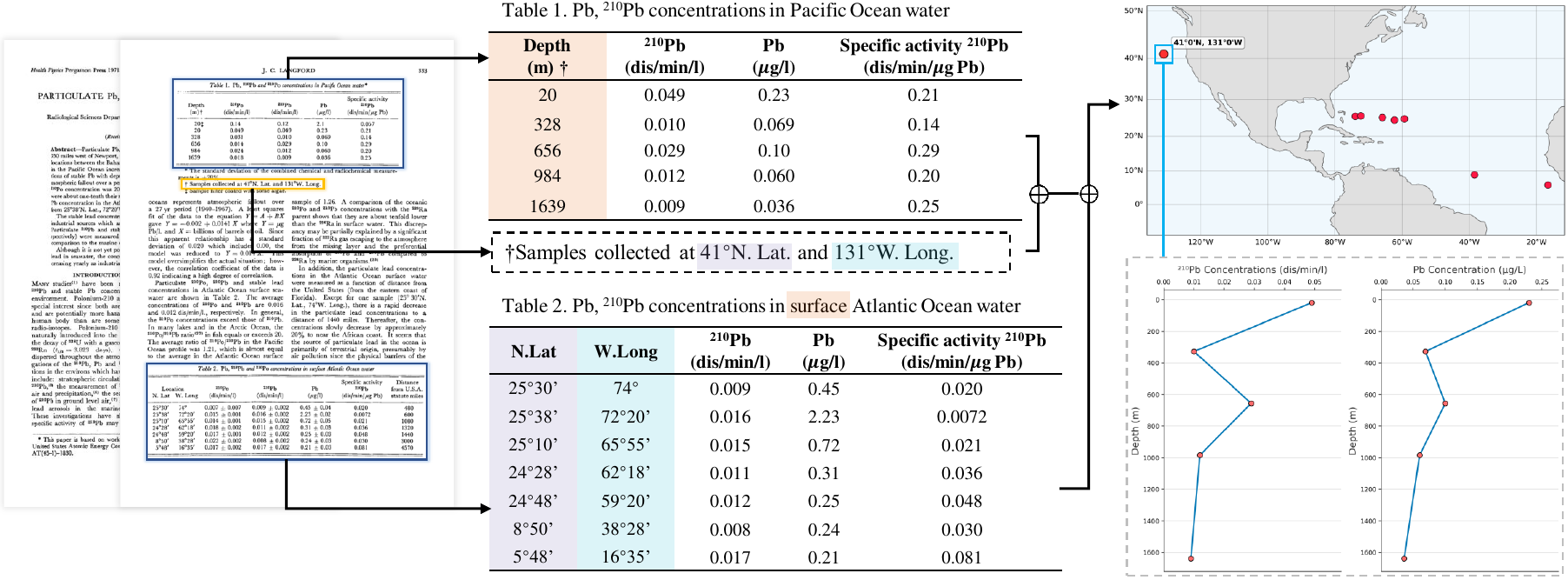}
  \caption{Case study of \textsc{Compass} data integration. \textsc{Compass} identifies relevant data tables, associates Pb measurements with metadata (coordinates from footnotes, depth from captions) within each table, and consolidates records into a unified dataset. Data sourced from Langford (1971)~\cite{Langford1971Pb}.}
  \Description{A composite figure showing a 1971 paper with two highlighted tables on the left, arrows extracting coordinates from footnotes and depth from captions, converging through join symbols to a map and two depth-profile line graphs on the right.}
  \label{fig:Pb_case}
\end{figure*}

\subsection{Deployment Results}

We deploy \textsc{Compass} with the Qwen2.5-32B backbone for full-scale marine Pb data integration, selected for its open-source nature and favorable performance-cost balance.

\subsubsection{Integration Overview}
\label{sec:deployment_overview}

We retrieve over 230,000 open-access papers using Pb-related keywords. \textsc{Compass} hierarchically processes these papers and identifies 110 containing target marine Pb data, extracting 3,751 new records. Combined with 12,704 records from 46 public datasets and 19,108 records from GEOTRACES~\cite{anderson2020geotraces}, the integration yields a total of 35,563 marine Pb records.

Compared to the 19,108 records in GEOTRACES—the most authoritative existing data product—our effort introduces 16,455 new records, representing an 86\% increase in available data. We categorize all records into eight types: Pb concentration, $^{210}$Pb concentration, and six isotope ratios ($^{206}$Pb/$^{204}$Pb, $^{207}$Pb/$^{204}$Pb, $^{208}$Pb/$^{204}$Pb, $^{206}$Pb/$^{207}$Pb, $^{208}$Pb/$^{206}$Pb, $^{208}$Pb/$^{207}$Pb). As shown in Figure~\ref{fig:Pb_map}, this significantly enhances global coverage, particularly in previously undersampled regions such as the East China Sea, Arabian Sea, and Southern Ocean. The complete source paper list is in Appendix~\ref{app:paper_list}. On benchmark papers containing target data, \textsc{Compass} achieves an average per-paper recall of 0.745, offering a practical proxy for within-paper extraction completeness.

The full pipeline required approximately 52 GPU hours on dual RTX 3090 GPUs, an efficiency improvement of over four orders of magnitude compared to manual curation~\cite{NUSSBAUMERSTREIT2021287}. The one-time Knowledge Tree construction cost is detailed in Section~\ref{sec:compass}.

\subsubsection{Data Quality Assessment}
\label{sec:data_quality}

\textbf{Automated Constraint Checks.}
All 3,751 extracted records are validated against physical constraints encoded in the Knowledge Tree's Validation Criteria. Value range verification and geographical boundary checks achieved 100\% pass rates, while unit conversion validation reached 95\%. Records that failed automated checks were flagged, rolled back, and re-processed by \textsc{Compass}. Rollback was triggered in approximately 0.06\% of processed cases across the full 230,000-paper corpus, with each instance limited to re-processing a single step, resulting in negligible additional computational overhead.

\textbf{Expert Manual Validation.}
We conducted stratified sampling across the 110 identified papers, reviewing 22 papers (20\%): 12 Pb conc., 4 $^{210}$Pb conc., 5 Pb isotopes ratios, and 1 mixed. These 22 papers yielded 946 data points, of which 869 were confirmed correct, giving 92\% accuracy (95\% CI: $\pm$1.7\%). The review was led by one marine scientist with a second providing overall consistency checks. Of the 77 erroneous data points ($\sim$8\%), errors stem primarily from semantic confusion (43, $\sim$56\%), e.g., rainwater samples collected during marine cruises misclassified as seawater, or sediment data confused with dissolved-phase measurements; followed by data association errors (26, $\sim$34\%), e.g., mismatched latitude/longitude or depth across tables; and PDF parsing failures (8, $\sim$10\%), e.g., row misalignment in upstream parsing.
Note that Table~\ref{tab:main_result} reports single-pass extraction for fair baseline comparison; with rollback enabled, \textsc{Compass} achieves F1: 0.619 
(see Table~\ref{tab:ablation}). The remaining gap with the 92\% deployment accuracy stems from two factors: the benchmark penalizes unextracted data regardless of paper relevance, and expert validation accepts minor coordinate variants (e.g., degree-minute conversions) that automated metrics penalize.

\subsubsection{Case Study}
Figure~\ref{fig:Pb_case} illustrates the \textsc{Compass} workflow on Langford (1971)~\cite{Langford1971Pb}. \textsc{Compass} identifies two relevant data tables and performs intra-table data association for each: in the first table, Pb measurements are linked with geographic coordinates recorded in its footnotes; in the second, data values are associated with the depth information specified in its caption. The independently resolved records from both tables are then consolidated into a unified, standardized dataset.

\begin{figure}[t]
  \centering
  \includegraphics[width=\linewidth]{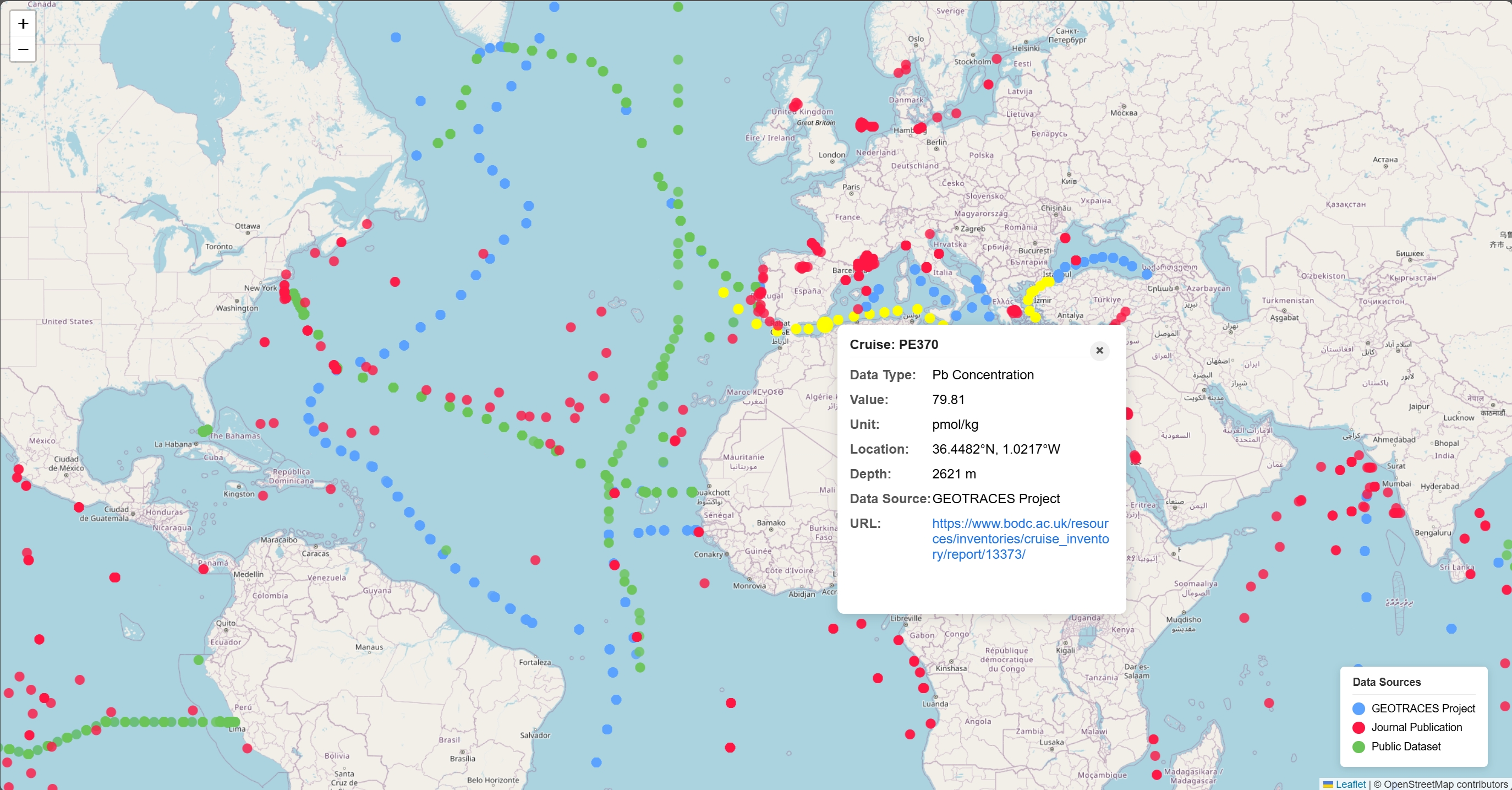}
  \caption{Screenshot of the online platform providing interactive spatial visualization, cruise-based data viewing, data querying, and source traceability.}
  \Description{A web platform screenshot with an interactive world map showing colored Pb data points across oceans. A popup card displays metadata for a selected North Atlantic record, including cruise ID, concentration value, location, depth, and source URL.}
  \label{fig:online_platform}
\end{figure}

\subsection{Online Data Platform}
We launch an interactive online platform\footnote{\url{https://jingwei.acemap.cn/lead}} integrated into the open-source JingWei marine data system for visualization and exploration of the Pb database. Users can query records with full metadata and source provenance (Figure~\ref{fig:online_platform}). Since launch, the platform has received over 1,590 visits, demonstrating its practical utility for marine scientific research.

\subsection{Scientific Insights}

The integrated marine Pb database provides a substantially enriched data landscape compared to existing resources, enabling scientific analyses previously infeasible due to data limitations.

\textbf{Enhanced Global Coverage.}
As shown in Figure~\ref{fig:Pb_map}, the integrated database extends spatial coverage across all eight Pb measurement types.
Among the 3,751 newly extracted records, concentration measurements constitute the largest share, with 1,228 Pb concentration and 472 $^{210}$Pb activity concentration records that directly support global Pb inventory estimation and vertical flux studies.
The three $^{204}$Pb-normalized isotope ratios ($^{206}$Pb/$^{204}$Pb: 562; $^{207}$Pb/$^{204}$Pb: 532; $^{208}$Pb/$^{204}$Pb: 476) provide the most diagnostic power for source apportionment, as $^{204}$Pb is the only non-radiogenic stable Pb isotope.
Additional paired ratios ($^{206}$Pb/$^{207}$Pb: 246; $^{208}$Pb/$^{206}$Pb: 139; $^{208}$Pb/$^{207}$Pb: 96), though fewer in number, complement source identification in regions where $^{204}$Pb measurements are unavailable.
This broader coverage reduces reliance on interpolation for global Pb budget estimation, particularly in the Southern Hemisphere and marginal seas.

\textbf{Regional Data Enrichment.}
The integration reveals previously under-sampled patterns in several key regions.
In the \textit{East China Sea}, newly incorporated Pb concentration and isotope data capture the influence of Yangtze River discharge and East Asian industrial emissions, filling a critical gap in global databases for this highly impacted marginal sea.
In the \textit{Southern Ocean}, recovered records from historical expedition reports provide valuable constraints on pre-industrial Pb baselines and the penetration of anthropogenic Pb into deep waters.
In the \textit{Arabian Sea}, new records complement GEOTRACES transects by adding temporal depth, enabling investigation of how regional Pb sources have evolved over decades.

\textbf{Implications for Anthropogenic Impact Assessment.}
The temporal span of the dataset—from the 1960s to the present—enables multi-decadal trend analysis. The global phase-out of leaded gasoline provides a natural experiment for studying oceanic recovery, and our data capture both the peak contamination and recovery phases across diverse regions. The comprehensive isotope coverage supports source apportionment, linking observed Pb patterns to anthropogenic emissions. These insights demonstrate that systematically recovering scattered historical records enables a more complete understanding of human impacts on ocean chemistry.

\section{Conclusion}
In this paper, we propose \textsc{Compass}, a Knowledge Tree-enhanced LLM Agent framework that operationalizes an expert-guided adaptation approach for fine-grained scientific data integration from academic papers. By co-designing a domain-specific Knowledge Tree with marine scientists, \textsc{Compass} bridges the gap between general-purpose LLMs and high-stakes scientific domains without fine-tuning. Deploying \textsc{Compass} across over 230,000 open-access papers, we recover 3,751 previously unincorporated Pb records, substantially expanding the integrated marine Pb database, with 92\% accuracy confirmed through expert validation. The newly integrated data reveal previously obscured patterns in under-sampled regions such as the East China Sea and the Southern Ocean. In the future, we plan to extend \textsc{Compass} to additional scientific domains and integrate multi-modal processing capabilities.

\section{Limitations and Ethical Considerations}
Despite \textsc{Compass}'s effectiveness, we acknowledge certain limitations delineating our current scope. 
Our implementation focuses on extracting data from tables and text, excluding figures. This is a strategic choice: the vast majority of marine Pb records are reported in tabular or textual format, whereas graphical representations are less commonly used for data reporting in this domain.
Furthermore, extraction quality is partially bounded by upstream PDF parsing tools. While layout recognition errors can propagate through the pipeline, our modular design ensures that \textsc{Compass} will naturally benefit from future advancements in general document analysis. 
Finally, the Knowledge Tree requires expert-guided initialization; while this incurs a setup cost, the construction process is iterative and accumulates reusable experience—including core module design and structural patterns—such that subsequent updates or generalization to other scientific domains primarily require domain-specific knowledge adaptation rather than rebuilding the entire framework from scratch.

\noindent\textbf{Ethical Considerations.} All data are sourced from publicly available open-access papers with no personally identifiable information. Our platform ensures transparency by linking every integrated record to its original source to support reproducibility.

\section{GenAI Disclosure}
During the preparation of this manuscript, generative AI tools were used to assist with language refinement and grammatical correction of certain draft passages. All scientific contributions, including methodology, experiments, analysis and conclusions, are solely the work of the authors.

\begin{acks}
This work was supported by National Natural Science Foundation of China (No. T2421002, 62602003, 62272293), Postdoctoral Fellowship Program of CPSF under Grant No. GZB20250806, and the AI for Science Seed Program of Shanghai Jiao Tong University (project number 2025AI4S-QY01).
\end{acks}

\bibliographystyle{ACM-Reference-Format}
\balance 
\bibliography{reference}

\appendix

\section{Complete Paper List}
\label{app:paper_list}
Table~\ref{tab:paper_list_part1} presents the complete list of 110 papers that were identified by \textsc{Compass} during its deployment in marine Pb data integration.

\begin{table*}[htbp]
    \centering
    \caption{Complete list of papers identified by \textsc{Compass} for marine Pb data integration.}
    \label{tab:paper_list_part1}
    \footnotesize
    \begin{tabular}{@{}r@{\hspace{3mm}}p{0.68\linewidth}@{\hspace{3mm}}p{0.21\linewidth}@{}}
        \toprule
        \textbf{No.} & \textbf{Title} & \textbf{DOI} \\
        \midrule
        1 & 210Pb and226Ra distributions in the Circumpolar waters & \textcolor{blue}{10.1016/0012-821x(81)90100-x} \\
        2 & 210Pb as a tracer of shelf–basin transport and sediment focusing in the Chukchi Sea & \textcolor{blue}{10.1016/j.dsr2.2008.10.021} \\
        3 & 210Pb sedimentation rates from the Northwestern Mediterranean margin & \textcolor{blue}{10.1016/j.margeo.2005.02.020} \\
        4 & 210Pb-derived chronology and the fluxes of 210Pb and 137Cs isotopes into continental shelf sediments, East Chukchi Sea, Alaskan Arctic & \textcolor{blue}{10.1016/0016-7037(95)00248-x} \\
        5 & 210Pb/226Ra and 210Po/210Pb disequilibria in seawater and suspended particulate matter & \textcolor{blue}{10.1016/0012-821x(76)90068-6} \\
        6 & 210Pb/226Ra disequilibrium in the Santa Barbara Basin & \textcolor{blue}{10.1016/0012-821x(75)90057-6} \\
        7 & 210Pb226Ra: Radioactive disequilibrium in the deep sea & \textcolor{blue}{10.1016/0012-821x(73)90194-5} \\
        8 & 226Ra and210Pb in the Weddell Sea & \textcolor{blue}{10.1016/0012-821x(80)90082-5} \\
        9 & 226Ra, 210Pb and 210Po disequilibria in the Western North Pacific & \textcolor{blue}{10.1016/0012-821x(76)90071-6} \\
        10 & 238U decay series nuclides in the northeastern Arabian Sea: Scavenging rates and cycling processes & \textcolor{blue}{10.1016/0278-4343(94)90015-9} \\
        11 & 210Pb and210Po in the equatorial Pacific and the Bering Sea: the effects of biological productivity and boundary scavenging & \textcolor{blue}{10.1016/s0967-0645(97)00024-6} \\
        12 & <sup>210</sup>Pb–<sup>226</sup>Ra–<sup>230</sup>Th systematics in very low sedimentation rate sediments from the Mendeleev Ridge (Arctic Ocean)This article is one of a series of papers published in this Special Issue on the theme<i> Polar Climate Stability Network</i>.GEOTOP Publication 2008-0031. & \textcolor{blue}{10.1139/e08-047} \\
        13 & A Voltammetric Study on Toxic Metals, their Speciation and Interaction with Nutrients and Organic Ligand in a South Pacific Ocean Region & \textcolor{blue}{10.1080/03067319808026847} \\
        14 & A comparison of the non-essential elements cadmium, mercury, and lead found in fish and sediment from Alaska and California & \textcolor{blue}{10.1016/j.scitotenv.2004.07.028} \\
        15 & A lead isotopic study of circum-antarctic manganese nodules & \textcolor{blue}{10.1016/0016-7037(95)00084-d} \\
        16 & Accumulation and potential sources of lead in marine organisms from coastal ecosystems of the Chilean Patagonia and Antarctic Peninsula area & \textcolor{blue}{10.1016/j.marpolbul.2019.01.026} \\
        17 & Accumulation of heavy metals (Pb, Zn, Cu, Cd), carbon and nitrogen in sediments from Strait of Georgia, B.C., Canada & \textcolor{blue}{10.1016/0304-4203(91)90017-q} \\
        18 & Analysis of 210Pb in sediment trap samples and sediments from the northern Arabian Sea: evidence for boundary scavenging & \textcolor{blue}{10.1016/s0967-0637(02)00013-4} \\
        19 & Antarctic Marine Sediments: Distribution of Elements and Textural Characters & \textcolor{blue}{10.1006/mchj.1998.1586} \\
        20 & Arctic vs. North Atlantic water mass exchanges in Fram Strait from Pb isotopes in sedimentsThis article is one of a series of papers published in this Special Issue on the theme <i>Polar Climate Stability Network</i>. & \textcolor{blue}{10.1139/e08-050} \\
        21 & Assessment of element concentrations in surface sediment samples from Sığacık Bay (eastern Aegean) & \textcolor{blue}{10.3906/yer-2002-15} \\
        22 & Basin-scale seawater lead isotopic character and its geological evolution indicated by Fe-Mn deposits in the SCS & \textcolor{blue}{10.1080/1064119x.2019.1637978} \\
        23 & Bioturbation in the abyssal Arabian Sea: influence of fauna and food supply & \textcolor{blue}{10.1016/s0967-0645(00)00052-7} \\
        24 & Brachidontes variabilis and Patella sp. as quantitative biological indicators for cadmium, lead and mercury in the Lebanese coastal waters & \textcolor{blue}{10.1016/j.envpol.2005.09.016} \\
        25 & Cadmium, copper and lead contamination of the seawater column on the Prestige shipwreck (NE Atlantic Ocean) & \textcolor{blue}{10.1016/s0003-2670(04)00333-2} \\
        26 & Cadmium, copper and lead contamination of the seawater column on the Prestige shipwreck (NE Atlantic Ocean) & \textcolor{blue}{10.1016/j.aca.2004.03.032} \\
        27 & Changes in sediment source areas to the Amerasia Basin, Arctic Ocean, over the past 5.5 million years based on radiogenic isotopes (Sr, Nd, Pb) of detritus from ferromanganese crusts & \textcolor{blue}{10.1016/j.margeo.2020.106280} \\
        28 & Colloid/Solution Partitioning of Metal-Selective Organic Ligands, and its Relevance to Cu, Pb and Cd Cycling in the Firth of Clyde & \textcolor{blue}{10.1006/ecss.1997.0267} \\
        29 & Comparative Base Line Studies on Pb-Levels in European Coastal Waters & \textcolor{blue}{10.1016/b978-0-08-022960-7.50017-8} \\
        30 & Comparative studies on trace metal levels in marine biota & \textcolor{blue}{10.1007/bf01359516} \\
        31 & Concentration and isotopic composition of dissolved Pb in surface waters of the modern global ocean & \textcolor{blue}{10.1016/j.gca.2018.05.005} \\
        32 & Concentration profiles of barium and lead in Atlantic waters off Bermuda & \textcolor{blue}{10.1016/0012-821x(66)90035-5} \\
        33 & Concentrations of Cu and Pb in the offshore and intertidal sediments of the west coast of Peninsular Malaysia & \textcolor{blue}{10.1016/s0160-4120(02)00073-9} \\
        34 & Datierung von Ostseesedimenten mit<sup>210</sup>Pb & \textcolor{blue}{10.1080/10256019008624336} \\
        35 & Deposition rates, mixing intensity and organic content in two contrasting submarine canyons & \textcolor{blue}{10.1016/j.pocean.2008.01.001} \\
        36 & Determination of lead isotope ratios in seawater by quadrupole inductively coupled plasma mass spectrometry after Mg(OH)2 co-precipitation & \textcolor{blue}{10.1016/s0584-8547(00)00176-2} \\
        37 & Determination of mass accumulation rates and sediment radionuclide inventories in the deep Black Sea & \textcolor{blue}{10.1016/0967-0637(94)90064-7} \\
        38 & Differing controls over the Cenozoic Pb and Nd isotope evolution of deepwater in the central North Pacific Ocean & \textcolor{blue}{10.1016/j.epsl.2004.12.009} \\
        39 & Dissolved trace metals in the surface waters of Puget Sound & \textcolor{blue}{10.1016/0025-326x(85)90568-5} \\
        40 & Distribution of dissolved and particulate 226Ra, 210Pb and 210Po in the Bismarck Sea and western equatorial Pacific Ocean & \textcolor{blue}{10.1071/mf99170} \\
        41 & Do decreased trace metal concentrations in surficial skagerrak sediments over the last 15–30 years indicate decreased pollution? & \textcolor{blue}{10.1016/0269-7491(94)90132-5} \\
        42 & Effects of bottom water dissolved oxygen variability on copper and lead fractionation in the sediments across the oxygen minimum zone, western continental margin of India & \textcolor{blue}{10.1016/j.scitotenv.2016.05.125} \\
        43 & Enrichment in Trace Metals (Al, Mn, Co, Cu, Mo, Cd, Fe, Zn, Pb and Hg) of Macro-Invertebrate Habitats at Hydrothermal Vents Along the Mid-Atlantic Ridge & \textcolor{blue}{10.1007/s10750-005-4758-1} \\
        44 & Evaluation of the use of common sculpin (Myoxocephalus scorpius) organ histology as bioindicator for element exposure in the fjord of the mining area Maarmorilik, West Greenland & \textcolor{blue}{10.1016/j.envres.2014.05.031} \\
        45 & Fe–Si-oxyhydroxide deposits at a slow-spreading centre with thickened oceanic crust: The Lilliput hydrothermal field (9°33'S, Mid-Atlantic Ridge) & \textcolor{blue}{10.1016/j.chemgeo.2010.09.012} \\
        46 & Geographic control on Pb isotope distribution and sources in Indian Ocean Fe-Mn deposits & \textcolor{blue}{10.1016/s0016-7037(01)00713-x} \\
        47 & Global environmental effects of large volcanic eruptions on ocean chemistry: Evidence from “hydrothermal” sediments (ODP Leg 185, Site 1149B) & \textcolor{blue}{10.1029/2007jb005333} \\
        48 & Gulf of Guinea continental slope and Congo (Zaire) deep-sea fan: Sr–Pb isotopic constraints on sediments provenance from ZaiAngo cores & \textcolor{blue}{10.1016/j.margeo.2005.11.014} \\
        49 & Heavy metal pollution and its relation to the malformation of green mussels cultured in Muara Kamal waters, Jakarta Bay, Indonesia & \textcolor{blue}{10.1016/j.marpolbul.2018.06.029} \\
        50 & Heavy metal sedimentation in Saanich Inlet measured with<sup>210</sup>Pb technique & \textcolor{blue}{10.1029/jc082i034p05477} \\
        51 & Heavy metals from Kueishantao shallow-sea hydrothermal vents, offshore northeast Taiwan & \textcolor{blue}{10.1016/j.jmarsys.2016.11.018} \\
        52 & High-precision measurements of seawater Pb isotope compositions by double spike thermal ionization mass spectrometry & \textcolor{blue}{10.1016/j.aca.2014.12.012} \\
        53 & Hydrothermal Fe–Si–Mn oxide deposits from the Central and South Valu Fa Ridge, Lau Basin & \textcolor{blue}{10.1016/j.apgeochem.2011.04.008} \\
        \bottomrule
    \end{tabular}
\end{table*}

\begin{table*}[htbp]
    \centering
    \caption*{\textbf{Table 4 (cont.)}}
    \label{tab:paper_list_part2}
    \footnotesize
    \begin{tabular}{@{}r@{\hspace{3mm}}p{0.68\linewidth}@{\hspace{3mm}}p{0.21\linewidth}@{}}
        \toprule
        \textbf{No.} & \textbf{Title} & \textbf{DOI} \\
        \midrule
        54 & Implications of excess 210Pb and 137Cs in sediment cores from Mikawa Bay, Japan & \textcolor{blue}{10.1016/s1001-0742(08)62328-1} \\
        55 & Intercomparison of alpha and gamma spectrometry techniques used in 210Pb geochronology & \textcolor{blue}{10.1016/j.jenvrad.2006.11.007} \\
        56 & In‐situ enrichment of heavy metals from deep‐sea water by an ion‐exchange pump system & \textcolor{blue}{10.1080/10641199609388319} \\
        57 & Isotopic analysis of metalliferous sediment from the East Pacific Rise & \textcolor{blue}{10.1016/0012-821x(71)90121-x} \\
        58 & Isotopic evidence for the source of lead in the North Pacific abyssal water & \textcolor{blue}{10.1016/j.gca.2010.05.017} \\
        59 & Isotopic tracing of anthropogenic Pb inventories and sedimentary fluxes in the Gulf of Lions (NW Mediterranean sea) & \textcolor{blue}{10.1016/s0278-4343(98)00070-3} \\
        60 & Late Quaternary variability of Mediterranean Outflow Water from radiogenic Nd and Pb isotopes & \textcolor{blue}{10.1016/j.quascirev.2010.06.021} \\
        61 & Lead in the western North Atlantic Ocean: Completed response to leaded gasoline phaseout & \textcolor{blue}{10.1016/s0016-7037(97)89711-6} \\
        62 & Lead in tropical marine systems: A review & \textcolor{blue}{10.1016/0048-9697(86)90071-9} \\
        63 & Lead isotopic composition of metalliferous sediments from the Nazca plate & \textcolor{blue}{10.1130/mem154-p199} \\
        64 & Lead isotopic disequilibria between plankton assemblages and surface waters reflect life cycle strategies of coastal populations within a northeast Pacific upwelling regime & \textcolor{blue}{10.4319/lo.1993.38.3.0670} \\
        65 & Lead-210 and polonium-210 in the surface water of the Pacific & \textcolor{blue}{10.2343/geochemj.5.165} \\
        66 & Levels of radionuclide concentrations in benthic invertebrate species from the Balearic Islands, Western Mediterranean, during 2012–2018 & \textcolor{blue}{10.1016/j.marpolbul.2019.110519} \\
        67 & Marine sediment contamination and dynamics at the mouth of a contaminated torrent: The case of the Gromolo Torrent (Sestri Levante, north-western Italy) & \textcolor{blue}{10.1016/j.marpolbul.2016.06.010} \\
        68 & Mineralogy, geochemistry, and Sr-Pb isotopic geochemistry of hydrothermal massive sulfides from the 15.2°S hydrothermal field, Mid-Atlantic Ridge & \textcolor{blue}{10.1016/j.jmarsys.2017.02.010} \\
        69 & Mixing of particles and organic constituents in sediments from the continental shelf and slope off Cape Cod: SEEP—I results & \textcolor{blue}{10.1016/0278-4343(88)90082-9} \\
        70 & Monthly variation of trace metals in North Sea sediments. From experimental data to modeling calculations & \textcolor{blue}{10.1016/j.marpolbul.2014.07.053} \\
        71 & Natural rates of sediment containment of PAH, PCB and metal inventories in Sydney Harbour, Nova Scotia & \textcolor{blue}{10.1016/j.scitotenv.2009.05.029} \\
        72 & Neoglacial change in deep water exchange and increase of sea-ice transport through eastern Fram Strait: evidence from radiogenic isotopes & \textcolor{blue}{10.1016/j.quascirev.2013.06.015} \\
        73 & New constraints on the Pb and Nd isotopic evolution of NE Atlantic water masses & \textcolor{blue}{10.1029/2007gc001766} \\
        74 & On the role of colloids in trace metal solid-solution partitioning in continental shelf waters: a comparison of model results and field data & \textcolor{blue}{10.1016/0278-4343(95)98840-7} \\
        75 & Particle fluxes and recent sediment accumulation on the Aquitanian margin of Bay of Biscay & \textcolor{blue}{10.1016/j.csr.2008.11.018} \\
        76 & Particle mixing rates in sediments of the eastern equatorial Pacific: Evidence from 210Pb, 239,240Pu and 137Cs distributions at MANOP sites & \textcolor{blue}{10.1016/0016-7037(85)90010-9} \\
        77 & Particulate Pb, 210Pb and 210Po in the Environment & \textcolor{blue}{10.1097/00004032-197103000-00011} \\
        78 & Particulate and dissolved 210Pb activities in the shelf and slope regions of the Gulf of Mexico waters & \textcolor{blue}{10.1016/s0278-4343(02)00017-1} \\
        79 & Patella vulgata, Mytilus minimus and Hyale prevosti as bioindicators for Pb and Se Enrichment in Alexandria coastal waters & \textcolor{blue}{10.1016/0025-326x(91)90184-t} \\
        80 & Pb isotope compositions of modern deep sea turbidites & \textcolor{blue}{10.1016/s0012-821x(00)00340-x} \\
        81 & Pb, Sr, and Nd isotopes in basalts and sulfides from the Juan de Fuca Ridge & \textcolor{blue}{10.1029/jb092ib11p11380} \\
        82 & Pb’s high sedimentation inside the bay mouth of Jiaozhou Bay & \textcolor{blue}{10.1088/1755-1315/100/1/012071} \\
        83 & Pollution history of heavy metals on the Portuguese shelf using 210Pb-geochronology & \textcolor{blue}{10.1016/j.scitotenv.2006.03.042} \\
        84 & Recent sedimentation and mass accumulation rates based on 210Pb along the Peru–Chile continental margin & \textcolor{blue}{10.1016/j.dsr2.2004.08.015} \\
        85 & Records of Holocene climatic fluctuations and anthropogenic lead input in elemental distribution and radiogenic isotopes (Nd and Pb) in sediments of the Gulf of Lions (Southern France) & \textcolor{blue}{10.1177/0959683619846973} \\
        86 & Residence times of surface water and particle-reactive 210Pb and 210Po in the East China and Yellow seas & \textcolor{blue}{10.1016/0016-7037(91)90305-o} \\
        87 & SWASV speciation of Cd, Pb and Cu for the determination of seawater contamination in the area of the Nicole shipwreck (Ancona coast, Central Adriatic Sea) & \textcolor{blue}{10.1016/j.marpolbul.2011.08.047} \\
        88 & Seawater quality assessment and identification of pollution sources along the central coastal area of Gabes Gulf (SE Tunisia): Evidence of industrial impact and implications for marine environment protection & \textcolor{blue}{10.1016/j.marpolbul.2017.12.012} \\
        89 & Sediment and organic carbon focusing in the Shelikof Strait, Alaska & \textcolor{blue}{10.1016/j.margeo.2005.06.036} \\
        90 & Sedimentary signals of the upwelling along the Zhejiang coast, China & \textcolor{blue}{10.1016/j.ecss.2019.106396} \\
        91 & Sedimentation of natural radionuclides on the seabed across the northern Japan Trench. & \textcolor{blue}{10.2343/geochemj.30.217} \\
        92 & Settling fluxes of U- and Th-series nuclides in the Bay of Bengal: results from time-series sediment trap studies & \textcolor{blue}{10.1016/s0967-0637(00)00016-9} \\
        93 & Some features of the trace metal biogeochemistry in the deep-sea hydrothermal vent fields (Menez Gwen, Rainbow, Broken Spur at the MAR and 9°50'N at the EPR): A synthesis & \textcolor{blue}{10.1016/j.jmarsys.2012.09.005} \\
        94 & Source and distribution of lead in the surface sediments from the South China Sea as derived from Pb isotopes & \textcolor{blue}{10.1016/j.marpolbul.2010.07.026} \\
        95 & Spatial and temporal distribution of Fe, Ni, Cu and Pb along 140°E in the Southern Ocean during austral summer 2001/02 & \textcolor{blue}{10.1016/j.marchem.2008.05.001} \\
        96 & Spatial and temporal distribution of heavy metals in coastal core sediments from the Red Sea, Saudi Arabia & \textcolor{blue}{10.1016/j.oceano.2017.03.003} \\
        97 & Strontium, lead and zinc isotopes in marine cores as tracers of sedimentary provenance: A case study around Taiwan orogen & \textcolor{blue}{10.1016/j.chemgeo.2007.10.024} \\
        98 & Sub-part per trillion levels of lead and isotopic profiles in a fjord, using an ultra-clean pumping system & \textcolor{blue}{10.1016/s0304-4203(99)00070-5} \\
        99 & Sulfur and lead isotopic compositions of massive sulfides from deep-sea hydrothermal systems: Implications for ore genesis and fluid circulation & \textcolor{blue}{10.1016/j.oregeorev.2016.10.014} \\
        100 & The behaviour of 210Pb and 226Ra in the eastern Irish Sea & \textcolor{blue}{10.1016/0265-931x(90)90025-q} \\
        101 & The distribution of lead concentrations and isotope compositions in the eastern Tropical Atlantic Ocean & \textcolor{blue}{10.1016/j.gca.2018.01.018} \\
        102 & The use of Pb-210 geochronology as a sedimentological tool: Application to the Washington continental shelf & \textcolor{blue}{10.1016/0025-3227(79)90039-2} \\
        103 & Trace elements in surface sediments from Kongsfjorden, Svalbard: occurrence, sources and bioavailability & \textcolor{blue}{10.1080/03067319.2017.1317762} \\
        104 & Trace metal variability, background levels and pollution status assessment in line with the water framework and Marine Strategy Framework EU Directives in the waters of a heavily impacted Mediterranean Gulf & \textcolor{blue}{10.1016/j.marpolbul.2014.07.054} \\
        105 & Trace metals (Co, Ni, Cu, Cd, and Pb) in the southern East/Japan Sea & \textcolor{blue}{10.1007/s12601-014-0006-9} \\
        106 & Trace metals in Antarctic copepods from the Weddell Sea (Antarctica) & \textcolor{blue}{10.1016/s0045-6535(02)00855-x} \\
        107 & Tracing the source of Pb using stable Pb isotope ratios in sediments of eastern Beibu Gulf, South China Sea & \textcolor{blue}{10.1016/j.marpolbul.2019.02.028} \\
        108 & Unusual ratios in the surface water of the Gulf of Lions & \textcolor{blue}{10.1016/s0399-1784(98)80030-3} \\
        109 & Vertical distribution of 210Pb and 226Ra and their activity ratio in marine sediment core of the East Malaysia coastal waters & \textcolor{blue}{10.1007/s10967-011-1206-8} \\
        110 & What controls the mixed‐layer depth in deep‐sea sediments? The importance of POC flux & \textcolor{blue}{10.4319/lo.2002.47.2.0418} \\
        \bottomrule
    \end{tabular}
\end{table*}

\end{document}